\pdfoutput=1

\documentclass[11pt]{article}

\usepackage[preprint]{acl}

\usepackage{times}
\usepackage{latexsym}

\usepackage[T1]{fontenc}

\usepackage[utf8]{inputenc}

\usepackage{microtype}

\usepackage{inconsolata}

\usepackage{booktabs}
\usepackage{graphicx}
\usepackage{orcidlink}
\usepackage{multirow}
\usepackage{makecell}
\usepackage{xcolor}
\usepackage{color}
\usepackage{tcolorbox}
\usepackage{colortbl}
\usepackage{pifont}
\usepackage{amsmath}
\usepackage{graphicx}
\usepackage{subcaption}
\usepackage{url}
\newcommand{\ignore}[1]{}
\newcommand{\ie}{\emph{i.e.,} }

\newcommand{\eg}{\emph{e.g.,} }
\newcommand{\etc}{\emph{etc}}
\definecolor{kellygreen}{rgb}{0.3, 0.73, 0.09}
\definecolor{alizarin}{rgb}{0.82, 0.1, 0.26}
\newcommand{\cmark}{{\color{kellygreen} \ding{51}}}
\newcommand{\xmark}{{\color{alizarin} \ding{55}}}
\renewcommand{\thefootnote}{\fnsymbol{footnote}}

\title{Towards Event-oriented Long Video Understanding}

\author{
\setcounter{footnote}{1}
	Yifan Du\textsuperscript{\rm{1}}\footnotemark[1],
	Kun Zhou\textsuperscript{\rm{2}}\footnotemark[1],
        \textbf{Yuqi Huo}\textsuperscript{\rm{4}},
        \textbf{Yifan Li}\textsuperscript{\rm{1}},
	\textbf{Wayne Xin Zhao}\textsuperscript{\rm{1}}\footnotemark[2],\\
        \textbf{Haoyu Lu}\textsuperscript{\rm{1}},
        \textbf{Zijia Zhao}\textsuperscript{\rm{3}},
        \textbf{Bingning Wang}\textsuperscript{\rm{4}},
        \textbf{Weipeng Chen}\textsuperscript{\rm{4}},
	\textbf{Ji-Rong Wen}\textsuperscript{\rm{1,2}} \\
        \textsuperscript{1}Gaoling School of Artificial Intelligence, Renmin University of China \\
	\textsuperscript{2}School of Information, Renmin University of China \\
        \textsuperscript{3}Institute of Automation, Chinese Academy of Sciences\\
        \textsuperscript{4}Baichuan Inc.\\
	 \texttt{\{yifandu1999,                           
              batmanfly\}@gmail.com}, \texttt{francis\_kun\_zhou@163.com} \\
}

\begin{document}
\maketitle
\footnotetext[1]{Equal contribution.}
\footnotetext[2]{Corresponding author.}
\renewcommand{\thefootnote}{\arabic{footnote}}
\begin{abstract}
With the rapid development of video Multimodal Large Language Models (MLLMs), numerous benchmarks have been proposed to assess their video understanding capability. However, due to the lack of rich events in the videos, these datasets may suffer from the short-cut bias that the answers can be deduced from a few frames, without the need to watch the entire video. To address this issue, we introduce \emph{\textbf{Event-Bench}}, an event-oriented long video understanding benchmark built on existing datasets and human annotations. Event-Bench includes six event-related tasks and 2,190 test instances to comprehensively evaluate video event understanding ability. Additionally, we propose \emph{\textbf{Video Instruction Merging~(VIM)}}, a cost-effective method that enhances video MLLMs using merged, event-intensive video instructions, addressing the scarcity of human-annotated, event-intensive data. Extensive experiments show that the best-performing model, GPT-4o, achieves an overall accuracy of 53.33, significantly outperforming the best open-source model by 41.42\%. Leveraging an effective instruction synthesis method and an adaptive model architecture, VIM surpasses both state-of-the-art open-source models and GPT-4V on the Event-Bench. All code, data, and models are publicly available at \url{https://github.com/RUCAIBox/Event-Bench}.

\end{abstract}
   
\section{Introduction}
\begin{figure}
    \centering
    \includegraphics[width=1\linewidth]{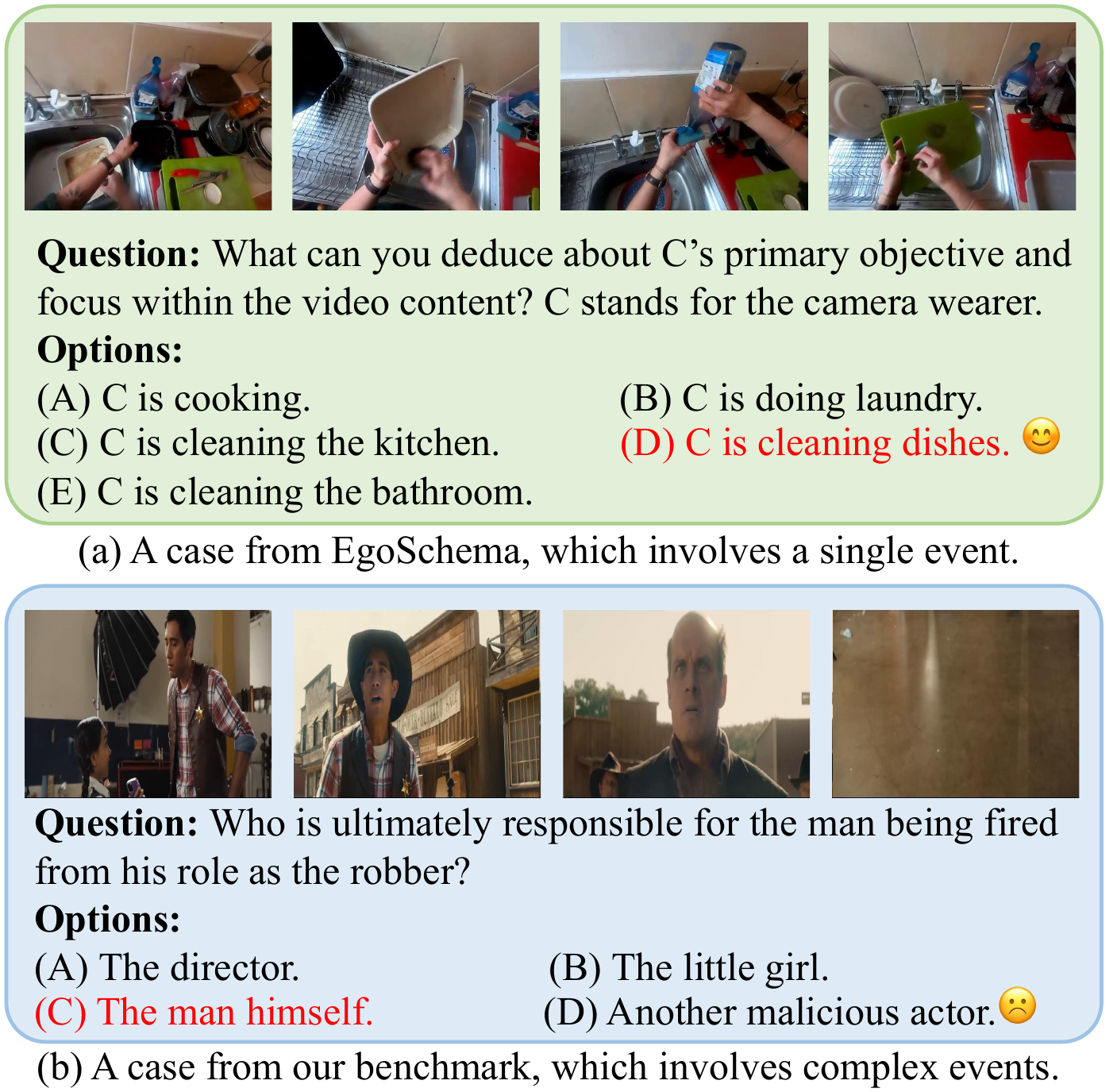}
    \caption{The comparison of two representative examples from existing benchmarks and our Event-Bench.}
    \label{fig:intro}
\end{figure}
Video understanding is a key capability for AI models to perceive the visual world like humans. It requires models to recognize features and changes in regions or objects and to comprehend the overall context and storyline throughout the video. Building upon Large Language Models~(LLMs)~\cite{brown-2020-nips-language, touvron-2023-arxiv-llama, Zhao-2023-arxiv-survey}, current Video Multimodal Large Language Models~(video MLLMs)~\cite{Tang-2023-Video, Zhang-2023-Video-LLaMA, Maaz-Video-ChatGPT-2023} exhibit surprising video understanding capabilities. Concurrently, numerous benchmarks are proposed to evaluate their performance in various video understanding scenarios, \eg contextual reasoning~\cite{Mangalam-2023-egoschema} and situated reasoning~\cite{Wu-2021-STAR}. 

Despite these advancements, recent work has found that these datasets may suffer from the short-cut bias~\cite{Lei-2023-revealing}. This bias refers to the phenomenon that answers to some questions can be deduced without fully watching the video, thereby affecting the evaluation's reliability. As illustrated in Figure~\ref{fig:intro}(a), although the video lasts for 3 minutes, it simply describes the behavior of cleaning dishes. Consequently, questions related to the video can be easily answered by viewing just a single frame. Essentially, the cause of the short-cut bias is the \emph{lack of rich events} in the videos. Events are the high-level semantic concepts that humans perceive when observing a video~\cite{Lavee-2009-understanding} (\eg the moment a player makes a shot in a soccer match), which are crucial for representing the unique and dynamic insights that differentiate various videos. Since the necessity of event-oriented video understanding might be neglected in existing datasets, their annotated test instances may fail to accurately estimate human-like video understanding capability.


In light of this, we present an event-oriented long video understanding benchmark, namely \emph{\textbf{Event-Bench}}. It focuses on comprehensively evaluating video MLLMs across three levels of event understanding capabilities: atomic, composite, and overall understanding, encompassing six event-related tasks. To construct it, we design an automatic pipeline to meticulously collect unbiased test instances for these tasks from existing datasets, unifying their formats and filtering out low-quality ones. Besides, we manually craft test instances based on event-intensive long videos from YouTube to cover complex real-world scenarios. In total, Event-Bench contains 2,190 samples. As shown in Table~\ref{tab:intro}, our benchmark distinguishes itself with longer time scopes and an event-oriented focus.

To elicit human-like video understanding capabilities, it is necessary to utilize massive event-intensive video instruction for training video MLLMs~\cite{chen2024sharegpt4video}, but annotating such data is costly. To address this, we leverage existing image and video instructions to compose more complex training data. Specifically, we first employ an adaptive model architecture to handle both image and video inputs, allowing us to incorporate high-quality image instructions into the training process. Second, we propose \emph{\textbf{Video Instruction Merging~(VIM)}}, which merges similar videos in the existing dataset into a new video containing all the events from the original videos. Extensive experiments on our Event-Bench demonstrate that our method outperforms all open-source models of comparable parameter scales and even surpasses GPT-4V on average (\ie 41.64 v.s. 32.65).


Our main contributions are listed as follows: 

(1) We propose an event-oriented long video benchmark, \textit{\textbf{Event-Bench}}, to evaluate the human-like video understanding capability;

(2) We devise \textbf{\textit{VIM}}, a low-cost method to improve video MLLMs using merged event-intensive video and high-quality image instructions;

(3) Experiment results show the comprehensive evaluation capability of Event-Bench for video MLLMs and the effectiveness of VIM.

\setlength{\tabcolsep}{3pt} 
\begin{table}
    \footnotesize
    \centering
    \begin{tabular}{lcccc}
    \toprule
         Benchmark&  \makecell[c]{Time\\Scope (s)}& \makecell[c]{Open\\Domain}&  \makecell[c]{Complex\\Reasoning}&  \makecell[c]{Event\\Oriented}\\
    \midrule
         MSVD-QA&  0$\sim$60& \cmark& \xmark& \xmark\\
         MSRVTT-QA&  10$\sim$30& \cmark& \xmark& \xmark\\
         TGIF-QA& - & \cmark& \xmark& \xmark\\
         ActivityNet-QA&  0$\sim$975& \xmark& \xmark& \xmark\\
         NExT-QA&  5$\sim$180& \cmark& \xmark& \xmark\\
         STAR&  2$\sim$195& \cmark& \cmark& \xmark\\
         CLEVRER&  5& \xmark& \cmark& \xmark\\
         EgoSchema&  180& \xmark& \cmark& \xmark\\
         MVBench&  5$\sim$40& \cmark& \cmark& \xmark\\
         TempCompass& 0$\sim$35&\cmark&\xmark&\xmark\\
         MovieChat& 401$\sim$602&\cmark&\xmark&\xmark\\
         Event-Bench& 2$\sim$1088&\cmark&\cmark&\cmark\\
    \bottomrule
    \end{tabular}
    \caption{Comparing our Event-Bench with existing video benchmarks, it is important to note that while videos in previous benchmarks may contain events, they are not specifically designed for event understanding. Event-Bench stands out due to its event-oriented design and longer time scope, making it uniquely suited for evaluating event comprehension. Further details can be found in the Appendix.}
    \label{tab:intro}
\end{table}


\section{Related Work}
\subsection{Video Multimodal Large Language Model}
Building upon the Large Language Model~(LLM), Multi-modal Large Language Models~(MLLMs) have recently obtained notable progress. Among them, video MLLMs exhibit surprising performance on various tasks~\cite{Zhang-2023-Video-LLaMA, Maaz-Video-ChatGPT-2023, Ren-2023-TimeChat}. \ignore{like video captioning~\cite{Zhang-2023-Video-LLaMA,Xu-2024-PLLaVA}, video question-answering~\cite{Maaz-Video-ChatGPT-2023,Lin-Video-LLaVA-2023,Li-2023-VideoChat}, and temporal video grounding~\cite{Ren-2023-TimeChat,Huang-2023-VTimeLLM}.} Typically, a video MLLM consists of a video encoder~(or image encoder), a LLM, and a connector to bridge these two components~\cite{Zhang-2023-Video-LLaMA,Li-2023-VideoChat,Maaz-Video-ChatGPT-2023}. Based on this type of architecture, the following works explore several ways to enhance the video MLLMs, \eg utilizing a more powerful video encoder~\cite{Lin-Video-LLaVA-2023}, supporting long context video~\cite{Song-2023-MovieChat,Wang-2024-LVCHAT}, and fine-tuning with large-scale instructions~\cite{Li-2023-MVBench}. In this work, we aim to synthesize video instructions with more complex events and explore scalable model architecture.

\ignore{However, since these models are evaluated on limited benchmarks, we are not sure of their performances on long video reasoning tasks, which is the most important perspective we should consider deploying a video MLLM in the real world. In this work, we propose a comprehensive benchmark to evaluate the reasoning ability of video MLLMs on long videos and develop a method to improve their performance from this perspective.}

\subsection{Video Understanding Benchmark}
Previous works propose benchmarks to evaluate various reasoning abilities in videos, including temporal reasoning~\cite{Xiao-2021-Nextqa}, situated reasoning~\cite{Wu-2021-STAR}, compositional reasoning~\cite{McLaughlin-2021-AGQA}, \etc. \ignore{STAR~\cite{STAR} introduces a benchmark that evaluates the situated reasoning ability, which requires the model to understand and perform reasoning on the surrounding environments. NextQA~\cite{} and CLEVRER aim to evaluate temporal reasoning and causal reasoning.} However, most videos in these benchmarks are short clips and lack diversity. With the development of video MLLMs, several works collect diverse videos to evaluate these models comprehensively~\cite{Ning-2023-videobench, Chen-2023-AutoEval}, but most videos in these benchmarks are no more than 1 minute. Following works like Egoschema~\cite{Mangalam-2023-egoschema} and MovieChat~\cite{Song-2023-MovieChat} collect long videos and create questions based on them. Despite this, the videos and questions in these benchmarks either do not involve complex reasoning in the event or are not open-domain. Therefore, we present an event-oriented long video understanding benchmark with diverse videos to comprehensively evaluate the model's ability to understand complex event narratives.

\ignore{VideoQA requires the model to accept a video as input and answer a question corresponding to the specific video. Existing VideoQA benchmarks mainly evaluate models from the perspective of perception~\cite{} and cognition~\cite{}. For video perception, most datasets focus on temporal understanding~\cite{tempcompass}, action recognition~\cite{activitynet-qa}, and object interaction~\cite{star}, \etc. For cognition reasoning, most datasets focus on temporal and causal reasoning~\cite{clevrer}, commonsense reasoning~\cite{nextqa}, and situated reasoning~\cite{star}, \etc. Although these benchmarks construct a comprehensive evaluation, they overlook a significant feature of video, events, which could divide videos into several levels according to the number of events they contain. Starting from this, we}

\section{Event-oriented Benchmark}

We propose Event-Bench, an event-oriented long video understanding benchmark for evaluating existing video MLLMs. It consists of massive videos, each paired with multi-choice questions from various event-related sub-tasks. To create this benchmark, we first establish a hierarchical task taxonomy and then collect data accordingly.

\subsection{Hierarchical Task Taxonomy}
\label{sec:task-taxonomy}

We organize our benchmark into three categories according to the number of events in a video, each of which comprises several sub-tasks.

\paragraph{Atomic Events Understanding.}
This task aims to evaluate the model's understanding of an atomic event (\eg an action of a human or object) in the video, which is one of the most basic video understanding capabilities.

\textbullet~\emph{Event Description.} 
For this sub-task, we collect question-answering pairs to evaluate whether the model can accurately recognize and describe a specific atomic event in the video, \eg ``What did the person do with the towel''.

\paragraph{Composite Events Understanding.} 
It focuses on understanding the relation between two atomic events in a video, from the following two aspects.

\textbullet~\emph{Temporal Reasoning.} 
We collect question-answer pairs that require reasoning about the temporal order of events in the video, \eg ``What did the man do after putting down the towel''.

\textbullet~\emph{Causal Reasoning.} 
This sub-task focuses on the causal relationship between two events in the video, particularly explaining why a specific event occurred, \eg ``Why did the man open the box''.
\paragraph{Overall Understanding.} 
It requires understanding the relationships across all events in a given video, to capture the high-level overall information from it.
We design the following three sub-tasks:

\textbullet~\emph{Contextual Reasoning.} 
This sub-task requires reasoning based on the overall context of the video, where the model needs to summarize content from a series of events, considering both actions and the environment, \eg ``Describe the overarching process conducted in the lab''.

\textbullet~\emph{Episodic Reasoning.} 
For a video, we also consider the episodes (\ie stories) involving characters and objects across all events. The model needs to understand high-level semantics to answer complex questions, \eg ``What led to Bean deciding to quickly leave the restaurant''. 

\textbullet~\emph{Counter-intuitive Reasoning.} 
For this sub-task, the videos involve counter-intuitive elements (\eg magical spells), and the model needs to identify the abnormal details to answer corresponding questions, \eg ``Why the video is magical''.


\subsection{Data Construction}
\label{sec:data-collection}
\begin{figure}[t]
    \centering
    \includegraphics[width=1\linewidth]{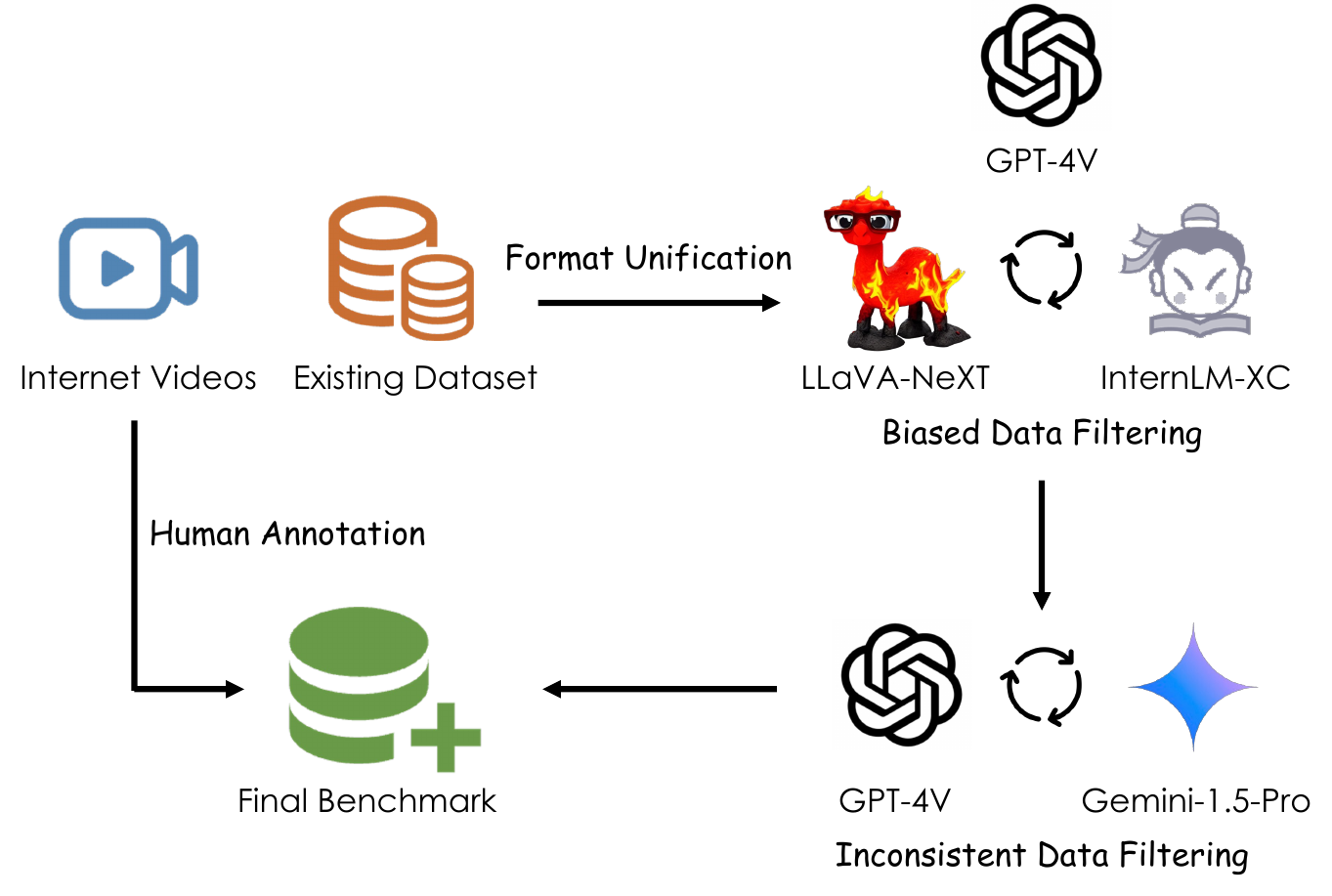}
    \caption{The data in Event-Bench are sourced from existing datasets or human annotations, involving three stages: format unification, biased data filtering, and inconsistent data filtering.}
    \label{fig:pipeline}
\end{figure}

\begin{figure*}[h]
    \centering
    \includegraphics[width=1\linewidth]{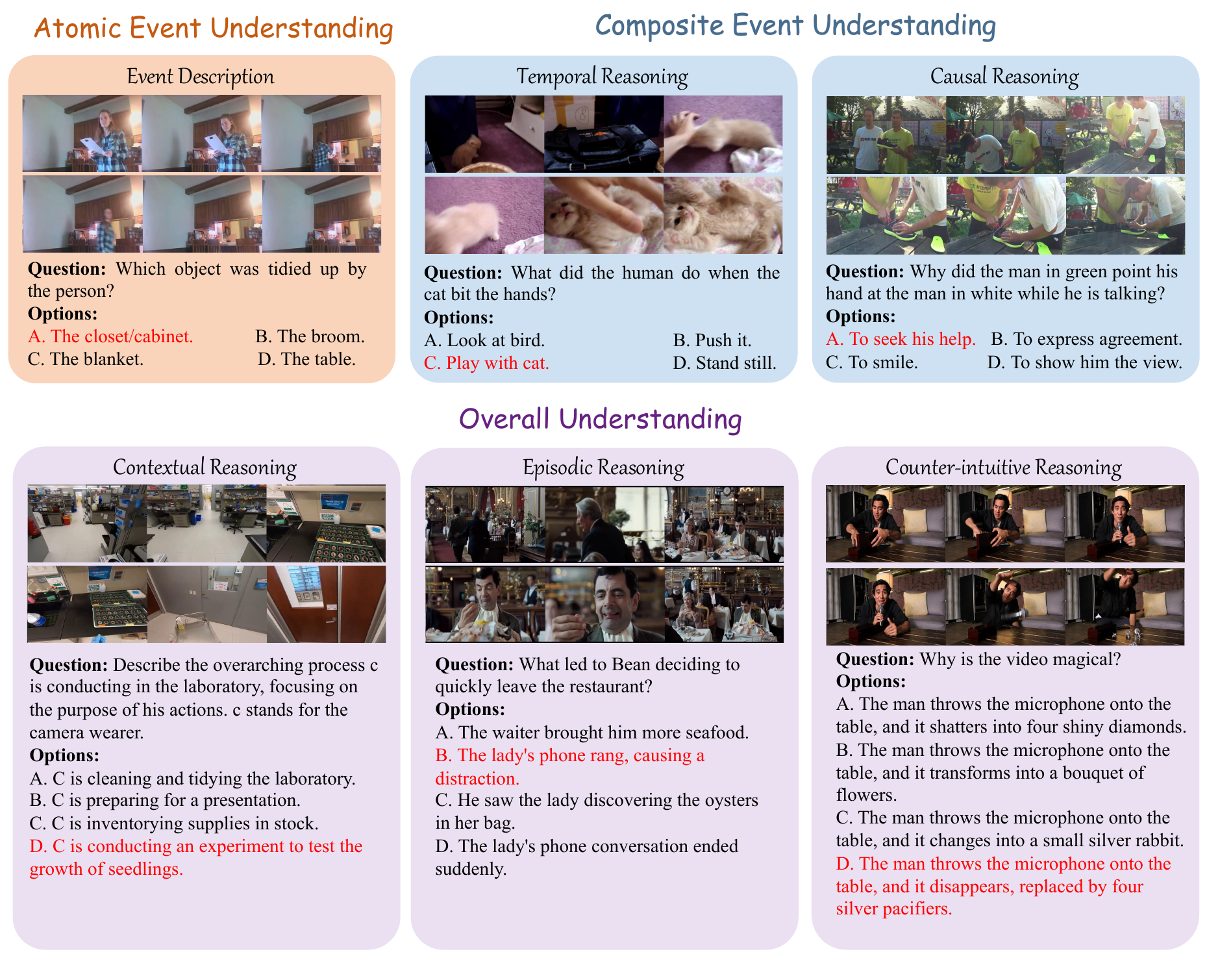}
    \caption{Overview of our Event-Bench. Our benchmark includes six sub-tasks across three event understanding abilities: atomic event understanding, composite event understanding, and overall understanding. The ground-truth answer is highlighted in red.}
    \label{fig:overview}
\end{figure*}

Our benchmark consists of data collected from existing datasets and newly human-annotated internet videos. The overall construction process is illustrated in Figure~\ref{fig:pipeline}.

\subsubsection{Construction Based on Existing Datasets}
Given the availability of multiple open-source VideoQA datasets, we aim to collect useful instances from them to create our event-oriented benchmark.
Specifically, we select instances from four datasets: STAR~\cite{Wu-2021-STAR}, NExT-QA~\cite{Xiao-2021-Nextqa}, EgoSchema~\cite{Mangalam-2023-egoschema}, and FunQA~\cite{Xie-2023-FunQA}, due to their diverse domains and rich annotations. However, after human review, we find three key issues in these instances: (1) different data formats and evaluation settings; (2) biased short-cut questions requiring no video understanding; (3) inconsistency between the answers and the video content. To address these issues, we develop the corresponding three-stage pipeline to preprocess the data.

\paragraph{Format Unification.} 
We first convert all open-ended questions into multi-choice questions using GPT-4, where the prompt is shown in the Appendix. The generated questions are further examined and revised by human annotators.

\paragraph{Biased Data Filtering.}
Inspired by existing work~\cite{Chen-2024-MMStar}, we filter the short-cut questions that can be answered using only a single frame of the video, as these represent biased test data for evaluating video understanding capabilities. Specifically, we employ three image MLLMs with different sizes and base LLMs(\ie GPT-4V~\cite{Openai-GPT-4V-2023}, LLaVA-NeXT-34B~\cite{liu2024llavanext}, and InternLM-XComposer2-4kHD~\cite{Dong-2024-xcomposer}) as the inspectors. If all these models can correctly answer a question using a single frame, we remove this sample from the dataset. This method effectively leverages shortcut bias to identify and eliminate biased data.


\paragraph{Inconsistent Data Filtering.} 
Finally, for each video and its corresponding question, we utilize two powerful MLLMs, \ie GPT-4V and Gemini-1.5-Pro\footnote{We sample 16 frames for GPT-4V and 1fps for Gemini-Pro-1.5 as the representation of the video.} to produce the answers. If their answers are the same but different from the human-annotated one, we regard the instance as an inconsistent sample and filter it out.

\subsubsection{Annotation Based on Internet Videos}

Although the processed instances from existing datasets are diverse and high-quality, we find that their videos generally contain relatively fewer events and their questions mostly neglect the episodic reasoning capability, which is important for testing the understanding capability of the overall video storyline. Therefore, we collect multiple videos from YouTube, which often feature user-generated content with diverse and narratively complex storylines, ensuring a rich set of events for the episodic reasoning task. We then annotate questions and answers specifically designed to test episodic reasoning. The questions constructed based on such videos are particularly challenging for the model. Considering the complexity of the episodic reasoning task, we decompose its annotation process into three stages: caption annotation, question generation, and answer check. 

\paragraph{Caption Annotation.} 
We ask human annotators to write the captions for every 30 seconds of a video. To ensure the quality, we first utilize Gemini-Pro-1.5 and GPT-4 to synthesize 10 questions per video, and ask human annotators to answer the questions based on their captions. This helps annotators to refine the captions and ensure they contain rich episodes for the following question generation stage. Although the synthetic questions may contain errors, they still guide the annotation process and help control the quality.

\paragraph{Question Generation.} 
To reduce human annotation costs, we utilize GPT-4 to generate the question-answer pairs for the episodic reasoning task based on the annotated captions. We provide the following prompt, along with detailed guidelines (in Appendix) to guarantee their consistency with the captions: \emph{``Based on the following descriptions, please ask 10 diverse questions about the plot and events of the video. While executing this task, please adhere to the following guidelines: ...''}

\paragraph{Answer Checking.}
We ask human annotators to answer the generated questions without seeing the corresponding answers generated by GPT-4. We then compare their answers for consistency. If the answers match, we add them to our benchmark. If not, we invite additional human annotators to review the question and vote on the final answer. Furthermore, we ask human annotators to select the time interval in the video that corresponds to the question-related event, which helps estimate annotation reliability.
\begin{table}[tb]
    \footnotesize
    \centering
    \begin{tabular}{c|cc|ccc|c}
    \toprule
         \textbf{Atomic}&  \multicolumn{2}{c|}{\textbf{Composite}}&  \multicolumn{3}{c|}{\textbf{Overall}}& \multirow{2}{*}{\textbf{Total}}\\
         ED&  TR&  CR&  CIR&  CU&  ER& \\
    \midrule
         468&  400&  400&  227&  395&  300& 2190\\
    \bottomrule
    \end{tabular}
    \caption{The statistic of Event-Bench. Each header is the abbreviation of the corresponding sub-tasks.}
    \label{tab:statistic}
\end{table}
\subsection{Data Statistics}
\label{sec:data-statistics}


Our benchmark consists of 2,190 video question-answer pairs across six tasks, assessing various event understanding abilities. Each task has 172 to 400 test samples. The hierarchical task taxonomy allows us to effectively evaluate models at different levels of capability. Additionally, since the benchmark is constructed from diverse data sources, it includes videos that cover a wide range of real-world domains and vary in length. These characteristics enable our benchmark to provide a comprehensive evaluation of existing video MLLMs. Examples from our benchmark are shown in Figure~\ref{fig:overview}.


\section{Methodology}

\begin{figure}[h]
    \centering
    \includegraphics[width=1\linewidth]{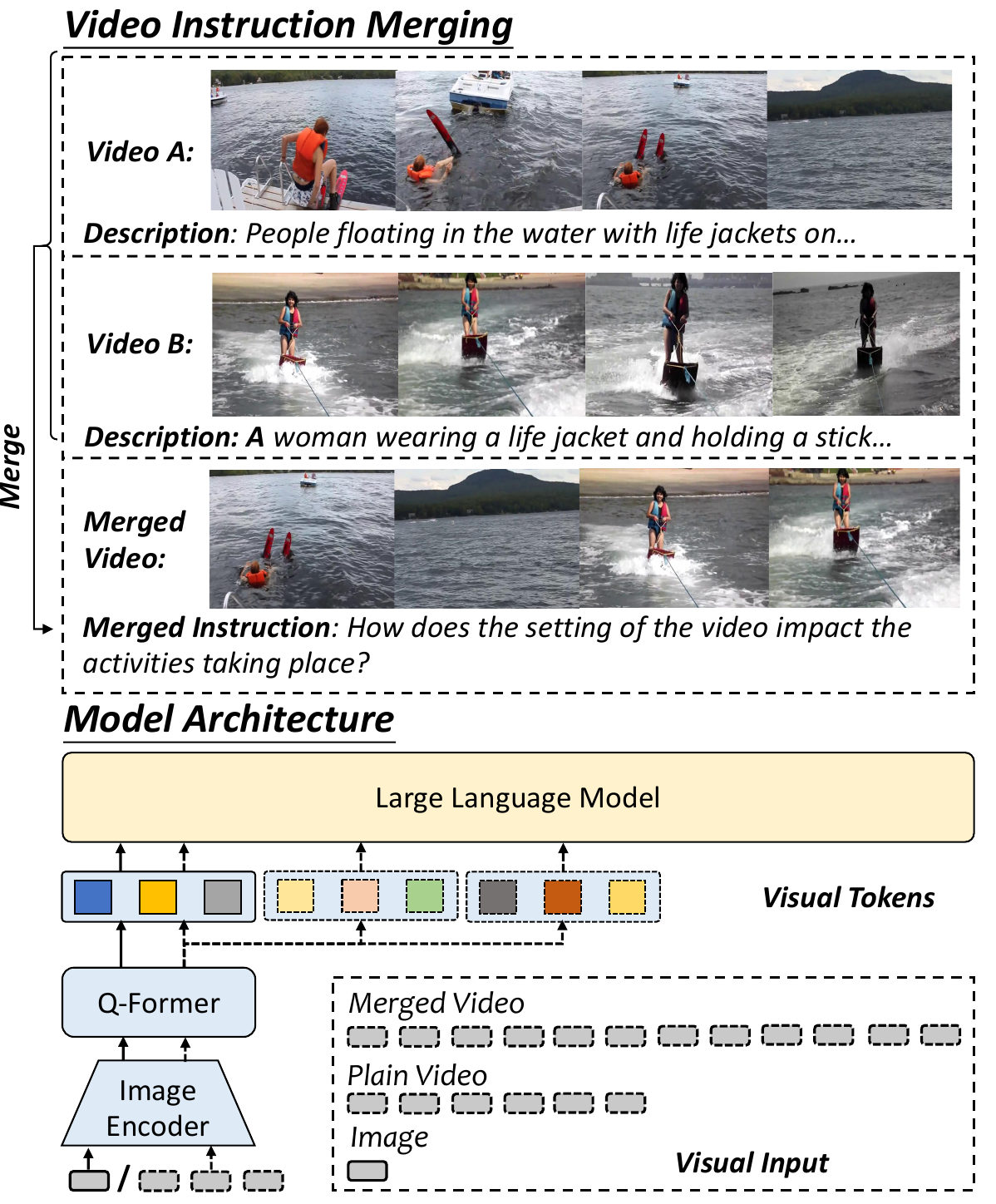}
    \caption{Overview of our method. We devise an instruction merging strategy to obtain instructions with more events based on existing data, and employ an adaptive model architecture supporting both image and video as the input.}
    \label{fig:model}
\end{figure}

In this section, we introduce Video Instruction Merging~(VIM) to enhance the performance of video MLLMs on event-oriented long video understanding tasks. Previous approaches primarily utilize video instruction tuning~\cite{Li-2023-VideoChat, Maaz-Video-ChatGPT-2023, Zhang-2023-Video-LLaMA} to improve the performance of video MLLMs on various tasks, typically requiring extensive human effort to annotate a large amount of video instructions. To address this, our proposed VIM integrates several similar video instructions from existing datasets into a new, event-intensive one as additional training data. We also adopt an adaptive model architecture in our video MLLM that interprets video as sequences of images, thereby handling both image and video inputs. This architecture allows us to combine existing high-quality image instructions with the newly created merged video instructions for training. The overall architecture of our approach is illustrated in Figure~\ref{fig:model}.

\subsection{Video Instruction Merging}
\label{sec:instruction}
Existing video instruction datasets suffer from the issues of lacking rich events~\cite{Heilbron-2015-activitynet}, \eg 1.41 per video on average for Video-ChatGPT-100K~\cite{Maaz-Video-ChatGPT-2023}. Thus, inspired by the mix-up strategy~\cite{Zhang-2018-mixup}, we propose to merge several simple video instructions into a single complex one with more events. Specifically, for each video and its corresponding instruction, we first find the most similar ones and then merge them into a new sample.

\paragraph{Similar Video Selection.}
We select the most similar video instructions to merge to ensure the coherence of the synthesized instructions. Specifically, we concatenate each question and answer into one sentence $[q_i;a_i]$, and convert it into the text embedding $\textbf{h}_i$ using the state-of-the-art BGE model~\cite{Chen-2024-BGE}. This embedding serves as the semantic representation of the entire instruction, and we compute its cosine similarity with other instructions to select the $k-1$ nearest neighbors:
\begin{equation}
    \text{Cos}(i,j) = \frac{\textbf{h}_i^{\top}\textbf{h}_j}{|\textbf{h}_i|*|\textbf{h}_j|}.
\end{equation}
In this way, we can divide the entire video instruction dataset $\mathcal{D}$ into $|\mathcal{D}|/k$ subsets.



\paragraph{Instruction Merging.}
For instructions within each similar video subset $\{v_i, q_i, a_i\}_{i=1}^{k}$, we merge them into a new one. We first temporally concatenate every video to create a new video $v'$. Then we ask ChatGPT~\footnote{https://chatgpt.com/} to generate a new question $q'$ and answer $a'$ for the merged video based on their original questions and answers. The process is formulated as follows:
\begin{equation}
\begin{aligned}
    v' &= [v_1 ; v_2 ; \dots ; v_k], \\ 
    q', a' &= \text{ChatGPT}(p_{m}, q_1, \dots, a_1, \dots),
\end{aligned}
\end{equation}
where $[;;]$ denotes the concatenation process and $p_m$ is the prompt for ChatGPT. We provide the complete prompt and some examples of merged data in the Appendix.

\begin{center}
\begin{tcolorbox}[colback=orange!5!white,colframe=orange!55!black,width=0.46\textwidth,title={Prompt for Instruction Merging}]
\small
The user will give you $k$ question-answer pairs about a video. These pairs have similar semantics but are different in some details. Your task is to create a new question-answer pair based on them, which requires the tester to watch all the videos to answer. The new question should be about the similarities and differences among these videos. The question should be diverse and the corresponding answer should be as detailed as possible...
\end{tcolorbox}
\end{center}



\subsection{Adaptive Model Architecture}
\label{sec:arch}
Our model architecture is composed of a scalable visual processor and an LLM. The scalable visual processor includes a reusable image encoder and a cross-modal connector. For video input, we first uniformly sample $n$ frames from it, then separately feed them into the visual processor to obtain the visual tokens. We then concatenate the resulting visual tokens to form the video representation. Image input is treated as in regular image MLLMs. Consequently, our model can flexibly handle inputs of varying sequence lengths (\eg a single image, short videos, or long videos). 

In practice, we employ a pre-trained Q-Former~\cite{Li-2023-BLIP2} as the cross-modal connector to reduce the number of visual tokens. These visual tokens are then concatenated with the embedding of question $q$ to serve as input to the LLM:
\begin{equation}
    \text{LLM}([\textbf{H}_{f_1}, \dots, \textbf{H}_{f_n}; \textbf{e}_1, \dots, \textbf{e}_L]),
\end{equation}
where $[\textbf{H}_{f_1}, \cdots, \textbf{H}_{f_n}]$ are the visual tokens and $[\textbf{e}_1, \textbf{e}_2, \cdots, \textbf{e}_L]$ are the text tokens. Since our model can handle both image and video inputs, we also include high-quality image instructions in our training data to help the LLM better align with and understand the visual input.

\section{Experiment}
\begin{table*}[h]
    \small
    \centering
    \begin{tabular}{l|c|ccc|cccc|c}
    \toprule
& \textbf{Atomic}& \multicolumn{3}{c|}{\textbf{Composite}}& \multicolumn{4}{c|}{\textbf{Overall}}&\multirow{3}{*}{\textbf{Avg.}}\\
\makecell[c]{Models}&  \makecell[c]{Event\\Description}&  \makecell[c]{Temporal\\Reasoning}&  \makecell[c]{Causal\\Reasoning}& \makecell[c]{Avg.}&  \makecell[c]{Counter\\Reasoning}&  \makecell[c]{Contextual\\Reasoning}&  \makecell[c]{Episodic\\Reasoning} &\makecell[c]{Avg.}& \\
    \midrule
    \multicolumn{10}{c}{\textit{Open-Source Image MLLMs}}\\
    \midrule
        LLaVA-NeXT (7B)&  13.68&  14.75&  9.75& 12.25&  14.98&  9.11&  7.30& 9.97 & 11.59\\
        IXC2-4KHD (7B)&  26.07&  27.50&  32.50& 30.00& 9.25&  12.15&  17.67 & 13.23 & 22.10\\
    \midrule
    \multicolumn{10}{c}{\textit{Open-Source Video MLLMs}}\\
    \midrule
        LLaMA-VID-long (7B)&  0.21&  0.00&  0.00& 0.00&  0.00&  0.00&  0.00 & 0.00 & 0.04\\
        LLaMA-VID (13B)&  1.92&  1.75&  0.00& 0.88&  3.08&  0.00&  4.00 & 2.06 & 1.60\\
        Video-LLaVA (7B)& 12.82& 5.50& 0.00& 2.75& 6.17& 2.78& 7.20 & 5.05 & 5.87\\
        Video-LLaMA (7B)&  15.81&  9.00&  6.25& 6.63&  0.09&  2.28&  0.67 & 1.22 & 6.68\\
        Video-ChatGPT (7B)*& 9.83& 9.50& 15.00& 12.25& 14.98& 12.66& 10.00& 12.37 & 11.78\\
        MovieChat (7B)*& 16.88& 16.00& 14.50& 15.25& 18.06& 13.16& \textbf{20.33}& 16.70 & 16.21\\
        PLLaVA (7B)& 34.62& 40.00& 40.50& 40.25& 17.62& 15.19& 11.00& 14.42 & 28.17\\
        VideoChat2 (7B)& 33.76& 37.75& 47.75& 42.75& 16.74& 15.70& 14.67 & 15.62 &29.41\\
        PLLaVA (13B)& 39.53& 42.50& 43.00& 42.75& \textbf{25.56}& 22.78& 17.00& 21.58 & 33.15\\
        ST-LLM (7B)&  47.22&  48.75&  59.50& 54.13&  9.69&  25.32&  16.67 & 18.66 & 37.71\\
        VIM (7B) (Ours)&  \textbf{48.08}&  \textbf{51.25}&  \textbf{61.25}& \textbf{56.25}&  22.91&  \textbf{32.66}&  18.67& \textbf{25.71} & \textbf{41.64}\\
    \midrule
    \multicolumn{10}{c}{\textit{Proprietary MLLMs}}\\
    \midrule
        GPT-4V& 29.70& 35.00& 40.00& 37.50& 36.56& 28.35& 27.00 & 29.93 & 32.65\\
        Gemini-1.5-Pro& 48.50& 47.50& 41.75& 44.63& 52.86& 32.15& \textbf{38.67}& 39.37 & 43.24\\
        GPT-4o&  \textbf{54.27}&  \textbf{56.75}&  \textbf{58.25}& \textbf{57.50}&  \textbf{63.44}&  \textbf{50.13}&  37.33 & \textbf{49.24} & \textbf{53.33}\\
    \bottomrule
    \end{tabular}
    \caption{Experiment results on Event-Bench. For the image MLLMs, we extract the frame in the middle of the video as the input. For the video MLLMs, we uniformly sample $\{8,16,32\}$ frames as the input and report the best performance. *Video-ChatGPT samples 100 frames, while MovieChat samples 1fps from the video.}
    \label{tab:main_exp}
\end{table*}

\subsection{Experimental Setup}
\paragraph{Implementation Details.} We utilize EVA-CLIP~\cite{Fang-2023-eva} as the image encoder, Vicuna-v1.1~\cite{vicuna2023} as the LLM, and initialize the Q-Former from InstructBLIP~\cite{Dai-2023-InstructBLIP}. We extend the maximum length of LLM from 2,048 to 4,096 tokens to accommodate inputs of up to 64 frames. For the training data, we utilize 100K instructions from Video-ChatGPT~\cite{Maaz-Video-ChatGPT-2023}, 40K instructions from Something-Something-2~\cite{Goyal-2017-ssv2}, 34K instructions from NExT-QA~\cite{Xiao-2021-Nextqa}, 10K from Vript~\cite{yang2024vript}, 100K image instructions from LLaVA665K~\cite{Liu-2023-improved}, and 32K instructions synthesized in Section~\ref{sec:instruction}. During training, we freeze the image encoder and the Q-Former, only updating the parameters of the LLM. We train our model on 8 Nvidia A100 (80G) GPUs for 1 epoch, completing the process within 12 hours.
\paragraph{Baseline Models.}
We select several SOTA MLLMs as the baselines. For open-source models, we select 2 image MLLMs~(LLaVA-NeXT~\cite{liu2024llavanext} and InternLM-XComposer2-4kHD~\cite{Dong-2024-xcomposer}) and 7 video MLLMs (Video-LLaMA, Video-ChatGPT~\cite{Maaz-Video-ChatGPT-2023}, MovieChat~\cite{Song-2023-MovieChat}, LLaMA-VID~\cite{Li-LLaMA-VID-2023}, VideoChat2~\cite{Li-2023-MVBench}, Video-LLaVA~\cite{Lin-Video-LLaVA-2023} and ST-LLM~\cite{Liu-ST-LLM-2024}). For proprietary models, we select GPT-4o, Gemini-1.5-Pro~\cite{Reid-Gemini1.5-2024}, and GPT-4V~\cite{Openai-GPT-4V-2023}.

\paragraph{Evaluation Protocols.}
We follow the evaluation strategy proposed in MMBench~\cite{Liu-MMBench-arxiv} to assess these models. We first use regular expressions to extract the options from the model's response. If successful, we use this as the prediction and compare it with the ground truth. Otherwise, we utilize GPT-4-turbo to determine if the prediction is correct. Besides, to ensure the consistency of models' responses on multiple-choice questions, we adopt the circular evaluation strategy~\cite{Liu-MMBench-arxiv}. Specifically, each question is fed to the model $N$ times~($N$ is the number of choices), where each time we shift the order of the choices. We consider the model to have succeeded in this sample if it provides the correct answer in every round.

\subsection{Main Results}

\begin{figure}
    \centering
    \includegraphics[width=1\linewidth]{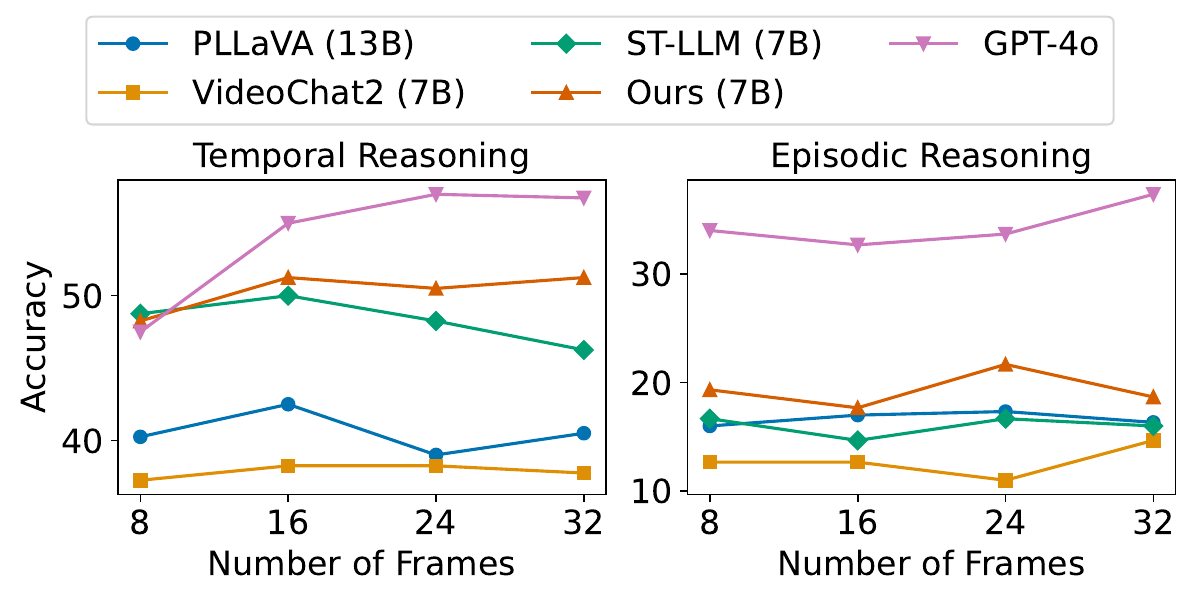}
    \caption{The relationship between the performance and the number of input frames.}
    \label{fig:acc_frame}
\end{figure}

The performance of the models is illustrated in Table~\ref{tab:main_exp}. We discuss the result and present the key findings from the following perspective:

\paragraph{Overall Performance.} As shown in Table~\ref{tab:main_exp}, both image MLLMs and video MLLMs exhibit poor performance on these event reasoning tasks. For the image MLLMs, LLaVA-NeXT and InternLM-XComposer2-4kHD could not achieve satisfactory performance when conditioned on only one frame, demonstrating the effectiveness of our data filtering strategies in building our benchmark. Surprisingly, most video MLLMs even underperform these two image MLLMs, implying their limited ability to understand complex events in videos. From a task perspective, we observe that overall understanding is more challenging than composite event understanding and atomic event understanding. Especially in our newly annotated episodic reasoning task, the most powerful Gemini-1.5-Pro and GPT-4o only achieve 38.67 and 37.33, respectively.

\paragraph{Comparisons of Different Models.} Most open-source models obtain comparable performance to proprietary models in atomic and composite understanding tasks, with some models even outperforming GPT-4V~(\eg ST-LLM, PLLaVA, and VideoChat2). However, the gap widens in the overall understanding task, where all the open-source models lag behind proprietary ones. Among the open-source models, our model performs the best across almost all tasks. The only exception is that MovieChat achieves the best on the episodic reasoning task and PLLaVA~(13B) is slightly better than ours on the counter-intuitive reasoning task. This is because MovieChat samples more frames and PLLaVA (13B) utilizes a larger LLM and more training data. Nevertheless, our model still achieves the highest average accuracy.

\subsection{Analysis}
\paragraph{Effect of Number of Frames.} Due to the limited context length in LLMs, most video MLLMs sample frames from the entire video uniformly as input. Intuitively, increasing the number of frames would help the model better understand the video, thus achieving better performance. We select the best four open-source models and one proprietary model to examine the relationship between their performance and the number of input frames, as shown in Figure~\ref{fig:acc_frame}. We can observe that more input frames lead to better performance for GPT-4o. For example, the performance of GPT-4o in the temporal reasoning task is boosted from 47.50 to 56.75 when the number of input frames increases from 8 to 32. However, open-source models do not consistently benefit from additional frames. Most achieve optimal performance with 16 or 24 frames, while increasing to 32 frames often leads to performance degradation. This indicates that developing a scalable model that could improve performance with increasing frame number is an important problem.

\paragraph{Effect of Training Strategy.}
We study the effect of the instruction merging strategy and the benefit of incorporating image data in our training process. First, the results in Table~\ref{tab:ablation} show that removing the merging strategy significantly hurts the performance across all tasks. Second, selecting videos with similar semantics leads to better performance than random selection, highlighting the importance of event coherence in videos. Regarding the effect of image data, we observe that removing image instructions from our training data leads to a performance decrease on all the tasks. This not only shows that image instruction could compensate for the lack of high-quality video data, but also demonstrates the compatibility and scalability of our model architecture.

\begin{table}
\small
    \centering
    \begin{tabular}{lcccc}
    \toprule
         &  Atomic&  Composite&  Overall& Avg.\\
    \midrule
         Ours&  \textbf{48.08}&  \textbf{56.25}&  \textbf{25.71}& \textbf{41.64}\\
         - w/o mixup&  43.16&  51.63&  24.39& 38.90\\
         - w/o image&  46.15&  51.75&  24.08& 38.90\\
         - random merge&  45.94&  54.25&  25.38& 40.32\\
    \bottomrule
    \end{tabular}
    \caption{Ablation study of VIM on Event-Bench.}
    \vspace{-5px}
    \label{tab:ablation}
\end{table}

\section{Conclusion}
In this work, we developed Event-Bench, an event-oriented benchmark for long video understanding, based on existing datasets and human annotations. Event-Bench comprises six event-related tasks and 2,190 test instances to comprehensively evaluate the capability of understanding events within videos. To address the lack of human-annotated, event-intensive video instructions, we devised an efficient training strategy to improve video MLLMs. We merged several simple video instructions into new, event-intensive ones, and we revised the model architecture to support the use of high-quality image instructions. Extensive experiments have shown that our Event-Bench provides a systematic comparison across different kinds of capabilities for existing video MLLMs, and points out the major shortcomings of open-source MLLMs. Besides, our approach outperforms all the open-source video MLLMs of comparable parameter scales and even surpasses GPT-4V on average.
\section{Limitation}
First, events are not only represented by visual modality, but also by other modalities in the real world (\eg textual, audio, and speech). These modalities convey important information in the video and complement the visual modality. As an initial exploration, Event-Bench only considers the visual modality and we plan to add other modalities to our benchmark in the future. Second, we only use 500K instructions during training the video MLLM due to the limited computational resources. However, experimental results show that including more high-quality video and image instructions positively impacts model performance. In the future, we aim to scale the training data and model size to achieve better performance. Third, although we synthesize video instructions in a cost-effective manner, their diversity and complexity are still lower than those of human-annotated ones. In the future, we will construct more event-intensive training data through human annotation.

\ignore{we develop a video MLLM that could receive 64 frames as the input, which is 4$\sim$8 times more than most existing open-source models. However, there is still a large gap between our model and the proprietary MLLMs in understanding super-long videos. Meanwhile, processing such long videos requires large computations and costs a lot of time. How to save the computation while maintaining the performance is a promising direction.}


\bibliography{custom}

\appendix
\clearpage
\setcounter{page}{1}
\section{Appendix}
\label{sec:appendix}
\subsection{Internet Videos Source}
We collect the internet videos from the following YouTube channels:

\url{https://www.youtube.com/@MrBean}

\url{https://www.youtube.com/@ZachKing}

\url{https://www.youtube.com/@filmmekker}

\url{https://www.youtube.com/@SimonsCat}
\subsection{Data Statistics} Our benchmark comprises a total of 2,190 video question-answer pairs on 6 tasks corresponding to different event understanding abilities, where each task has 172-400 test samples for evaluation.
\begin{figure}[h]
    \centering
    \includegraphics[width=0.8\linewidth]{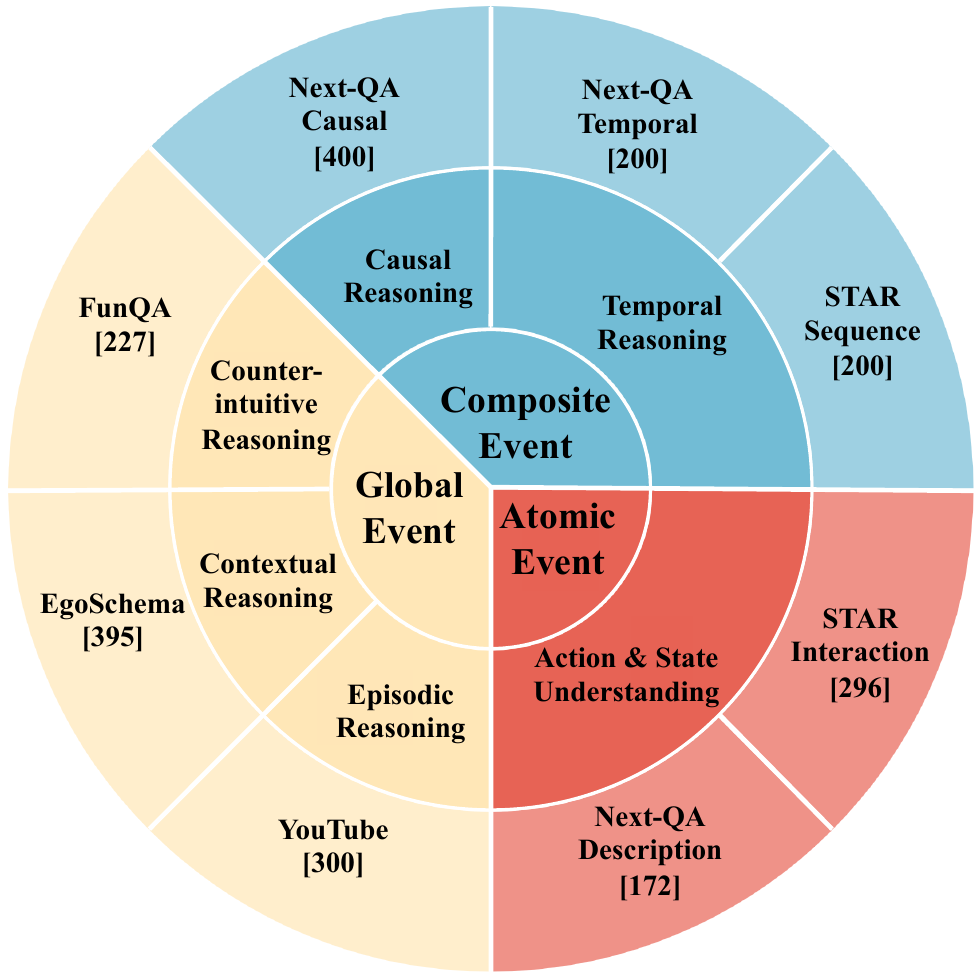}
    \caption{The dataset distribution of our benchmark.}
    \label{fig:statistic}
\end{figure}

\subsection{Ablation Study}
\begin{figure}[h]
    \centering
    \includegraphics[width=1\linewidth]{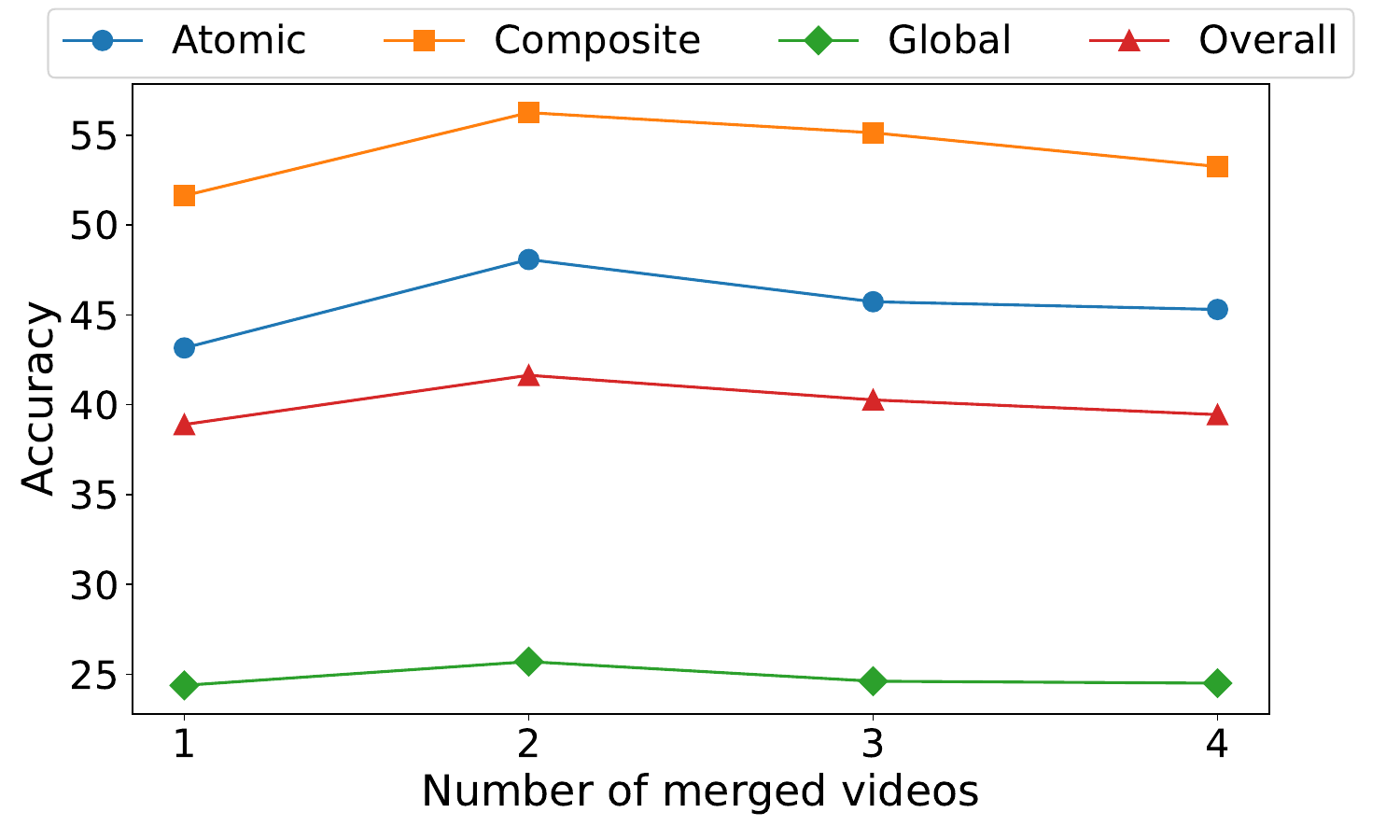}
    \caption{Performance comparison w.r.t the number of selected videos during video instruction merging.}
    \label{fig:merge_number}
\end{figure}

\paragraph{Number of Merged Videos.} In Section~\ref{sec:instruction}, we select $k$ samples and merge them into a new one, where a larger $k$ indicates more events happening in the new video. We experiment with $k=\{1,2,3,4\}$~($k=1$ indicates no merge operation) and depict the corresponding performance in Figure~\ref{fig:merge_number}. We could observe that increasing the number of events from 2 to 3 and 4 hurts performance on all the tasks, but is still better than the model trained on a single video. 

\begin{table*}[t]
    \small
    \centering
    \begin{tabular}{lcccccc}
    \toprule
         Benchmark&  Time Scope (s)&  Annotation& \makecell[c]{Open\\Domain}&  \makecell[c]{Complex\\Reasoning}&  \makecell[c]{Hierarchical\\Events}& \makecell[c]{Multiple\\Scenes}\\
    \midrule
         MSVD-QA~\cite{Xu-2017-msvd}&  0$\sim$60&  Auto& \cmark& \xmark& \xmark& \xmark\\
         MSRVTT-QA~\cite{Xu-2017-msvd}&  10$\sim$30&  Auto& \cmark& \xmark& \xmark& \xmark\\
         TGIF-QA~\cite{Jang-2017-TGIF}& - &  Auto+Human& \cmark& \xmark& \xmark& \xmark\\
         ActivityNet-QA~\cite{Yu-2019-activitynet}&  0$\sim$975&  Human& \xmark& \xmark& \xmark& \xmark\\
         NeXT-QA~\cite{Xiao-2021-Nextqa}&  5$\sim$180&  Human& \cmark& \xmark& \xmark& \xmark\\
         STAR~\cite{Wu-2021-STAR}&  2$\sim$195&  Auto& \cmark& \cmark& \xmark& \xmark\\
         CLEVRER~\cite{Yi-2020-CLEVRER}&  5&  Auto& \xmark& \cmark& \xmark& \xmark\\
         EgoSchema~\cite{Mangalam-2023-egoschema}&  180&  Auto& \xmark& \cmark& \xmark& \xmark\\
         MVBench~\cite{Li-2023-MVBench}&  5$\sim$40&  Auto& \cmark& \cmark& \xmark& \xmark\\
         TempCompass~\cite{Liu-2024-TempCompass}& 0$\sim$35& Auto+Human&\cmark&\xmark&\xmark&\xmark\\
         MovieChat~\cite{Song-2023-MovieChat}& 401$\sim$602& Human&\cmark&\xmark&\xmark&\cmark\\
         Ours& 2$\sim$1088& Auto+Human&\cmark&\cmark&\cmark&\cmark\\
    \bottomrule
    \end{tabular}
    \caption{Comparison with previous video understanding benchmarks.}
    \label{tab:intro-detail}
\end{table*}

\begin{table*}[h]
\footnotesize
    \centering
    \begin{tabular}{llc|cc|ccc}
    \toprule
 & & Atomic&\multicolumn{2}{c|}{Composite}&\multicolumn{3}{c}{Overall}\\
 \midrule
 \# Frames& \makecell[c]{Models}& \makecell[c]{Event\\Description}& \makecell[c]{Temporal\\Reasoning}& \makecell[c]{Causal\\Reasoning}& \makecell[c]{Counter-intuitive\\Reasoning}& \makecell[c]{Contextual\\Reasoning}&\makecell[c]{Episodic\\Reasoning} \\
 \midrule
          1 frame&LLaVA-NeXT (7B)&  13.68&  14.75&  9.75&  14.98&  9.11& 7.30 \\
 & IXC2-4KHD (7B)& 26.07& 27.50& 32.50& 9.25& 12.15&17.67 \\
 \midrule
 8 frame& LLaMA-VID (7B)& 0.00& 0.00& 0.00& 0.00& 0.00&0.00 \\
 & LLaMA-VID (13B)& 1.92& 1.75& 0.00& 3.08& 0.00&4.00 \\
 & Video-LLaVA (7B)& 12.82& 5.50& 0.00& 6.17& 2.78&7.20 \\
 & Video-LLaMA2 (7B)& 15.81& 9.00& 6.25& 0.09& 2.28&0.67 \\
 & VideoChat2 (7B)& 31.20& 37.25& 47.25& 14.98& 15.44&12.67 \\
 & ST-LLM (7B)& 47.22& 48.75& 59.50& 9.69& 25.32&16.67 \\
 & GPT-4V& 29.27& 32.75& 41.25& 42.29& 24.81&24.00 \\
 & GPT-4o& 48.08& 47.50& 55.50& 63.00& 48.86&34.00 \\
 \midrule
 16 frame& LLaMA-VID (7B)& 0.21& 0.00& 0.00& 0.00& 0.00&0.00 \\
 & LLaMA-VID (13B)& 1.06& 1.13& 0.25& 3.08& 0.00&5.00 \\
 & Video-LLaMA2 (7B)& 11.11& 3.25& 6.00& 0.88& 3.04&0.33 \\
 & PLLaVA& 34.62& 40.00& 40.50& 17.62& 15.19&11.00 \\
 & VideoChat2 (7B)& 34.19& 38.25& 46.25& 17.18& 17.22&12.67 \\
 & ST-LLM (7B)& 47.65& 50.00& 56.50& 11.45& 26.84&14.67 \\
 & GPT-4V& 29.70& 35.00& 40.00& 36.56& 28.35&27.00 \\
 & GPT-4o& 52.99& 55.00& 58.25& 63.00& 49.11&32.67 \\
 \midrule
 32 frame& LLaMA-VID (7B)& 0.00& 0.00& 0.00& 0.00& 0.00&0.00 \\
 & LLaMA-VID (13B)& 0.85& 0.75& 0.00& 3.08& 0.00&3.00 \\
 & Video-LLaMA2 (7B)& 9.19& 4.75& 3.75& 2.20& 1.77&1.33 \\
 & VideoChat2 (7B)& 33.76& 37.75& 47.75& 16.74& 15.70&14.67 \\
 & ST-LLM (7B)& 46.79 & 46.25& 55.25& 10.13& 26.33&16.00 \\
 & GPT-4V& 23.72& 25.75& 33.00& 40.09& 20.51&20.67 \\
 & GPT-4o& 54.27& 56.75& 58.25& 63.44& 50.13&37.33 \\
 \midrule
 more frames& MovieChat (7B)& 16.88& 16.00& 14.50& 18.06& 13.16& 20.33\\
 & Video-ChatGPT (7B)& 9.83& 9.50& 15.00& 14.98& 12.66&10.00 \\
          &Gemini-1.5-Pro&  48.50&  47.50&  41.75&  52.86&  32.15& 38.67 \\
    \bottomrule
    \end{tabular}
    \caption{Detailed experimental results with more frames as input.}
    \label{tab:main_result-detail}
\end{table*}
\vspace{-0.5cm}
\subsection{Prompt}

Among all the open-source datasets we used, only the HumorQA and MagicQA in FunQA dataset~\cite{Xie-2023-FunQA} do not have multiple-choice questions. We utilize the following prompt to convert it to multi-choice questions by GPT-4:
\begin{center}
\begin{tcolorbox}[colback=orange!5!white,colframe=orange!55!black,width=0.48\textwidth,title={Prompt for multi-choice question conversion.}]
\small
There is a task that requires finding a humorous/magical moment in a video, and several people write the descriptions of the video. Please change this task into a 4-way multi-choice question based on their descriptions. Ensure that:

1. The question must be 'Why the video is humorous/magical?' or 'What is the humorous/magical moment of the video?'

2. All the options you provide should be roughly the same length. The incorrect options should looks very humorous/magical and should NOT deviate too much from the description, but express different meanings with the correct option.
3. The question needs to involve details in the video, ensuring that only those who have watched the complete video can answer, and those who have only watched a small section cannot answer correctly.

Please provide the question in the following format:

Question: <question>

Options:

(A) xxx

(B) xxx

(C) xxx

(D) xxx

Let’s begin this task.

\#\#Video Description\#\#
\end{tcolorbox}
\end{center}

\begin{center}
\begin{tcolorbox}[colback=orange!5!white,colframe=orange!55!black,width=0.48\textwidth,title={Prompt for merging video instructions.}]
\small
You are a helpful human assistant who is helping a user with a task.

------

\#\#TASK:

The user will give you 5 questions and 5 answers about a video. These questions (answers) have very similar semantics but actually express different meanings, and their corresponding videos are also different.
Your task is to create a new question-answer pair based on these 5 question-answer pairs, and the tester must watch all the videos to answer. The created question should be about the similarities and differences among these videos. The question should be diverse and the corresponding answer should be as detailed as possible.

------

\#\#INSTRUCTION:

1. Do not ask questions that can be answered by watching only one video.

2. Do not use words like "as mentioned in the description" in both questions and answers.

3. Please give your response in the following format:

Question: 

Answer:

------

Let’s begin this task.

Question1: {question1}

Answer1: {answer1}

Question2: {question2}

Answer2: {answer2}
\end{tcolorbox}
\end{center}

\begin{center}
\begin{tcolorbox}[colback=orange!5!white,colframe=orange!55!black,width=0.48\textwidth,title={Prompt for episodic reasoning question generation.}]
\small
As an AI visual assistant, your task involves first analysing video content, and then ask questions about the video. Below are descriptions of each scene in a video, arranged in chronological order. Based on these descriptions, please ask 10 diverse questions about the plot and events of the video.

While executing this task, please adhere to the following guidelines:

1. These 10 questions must be very difficult, and only those who have watched the entire video can answer them correctly. Those who have only watched a few frames of the video cannot answer them correctly.

2. The questions need to be in the form of four-way multiple-choice questions, with only one option is the correct answer while the others are incorrect answers.

3. All the options you provide should be roughly the same length. The choices you present should be formulated in a way that makes them tricky to differentiate, thus the tester cannot guess the correct answer through common sense. They should only be able to answer the question by watching the entire video.

4. The description may involve some scenes where the director makes the video. Please ignore this part and do not ask questions about it.

5. Please give the rationale for the correct answer in the question, so that the tester can understand why the answer is correct. The answer must be based on the event happening in the video but not hypothesis.
Please provide 10 questions in the following format:

Question: <question>

(A) xxx

(B) xxx

(C) xxx

(D) xxx

Answer: <answer>

Rationale: <rationale>

Let’s begin this task.

\#\#Video Description\#\#
\end{tcolorbox}
\end{center}

\subsection{Video Instruction Merging}
\begin{figure*}[t]
    \centering
    \includegraphics[width=1\linewidth]{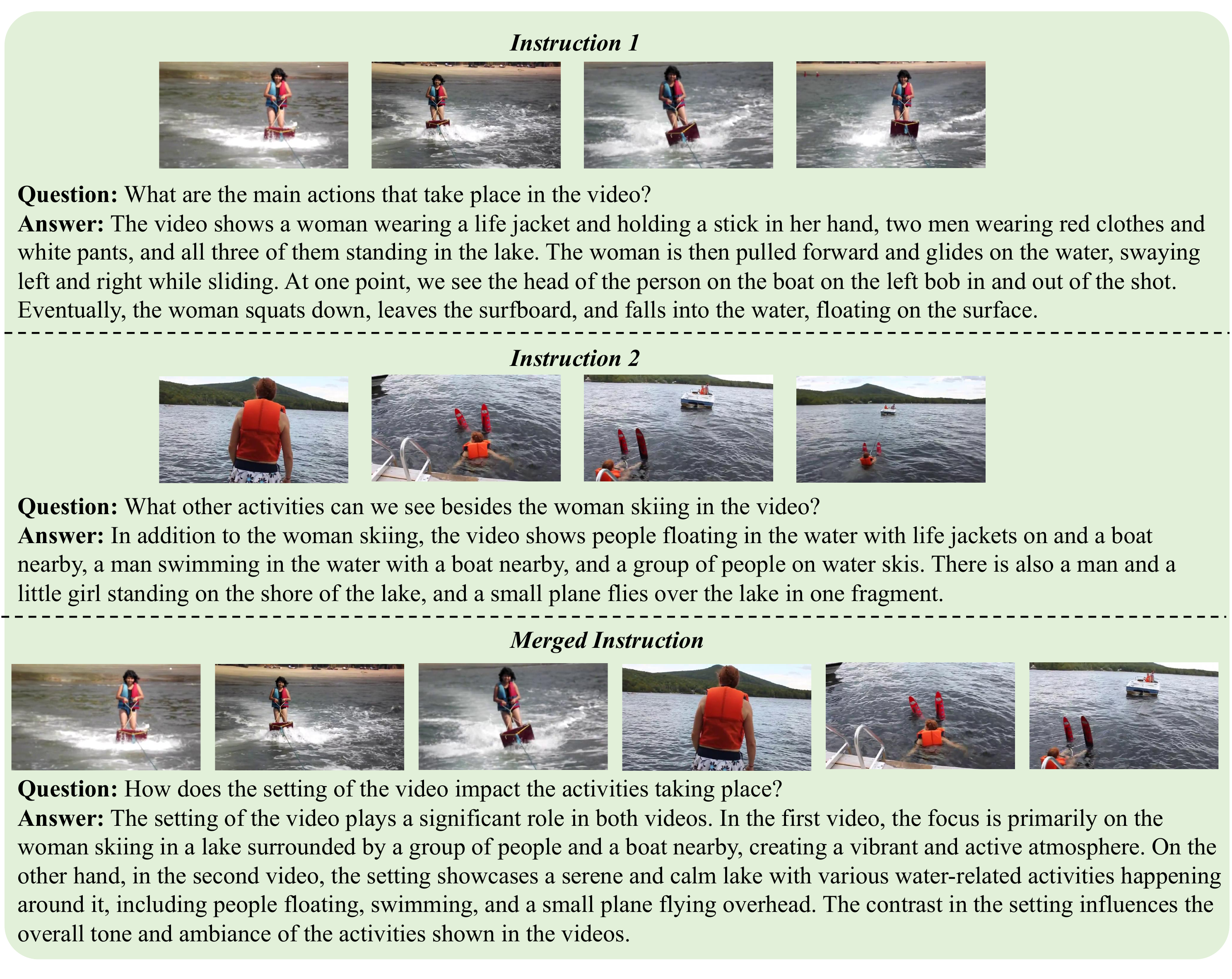}
    \caption{An example of the merged video instruction. \textbf{\textit{Instruction 1}} and \textit{\textbf{Instruction 2}} are two similar video instructions before merging, while \textit{\textbf{Merged Instruction}} is the data after merging.}
    \label{fig:vim}
\end{figure*}

We display an example displaying the process of video instruction merging in Figure~\ref{fig:vim}.

\end{document}